\documentclass[letterpaper]{article} 
\usepackage[preprint]{aaai2027}  
\usepackage[hyphens]{url}  
\usepackage{graphicx} 
\usepackage{natbib}  
\usepackage{caption} 
\usepackage{algorithm}
\usepackage{algorithmic}
\usepackage{url}            
\usepackage{booktabs}       
\usepackage{graphicx}
\usepackage{amsmath}
\usepackage{amsfonts}       
\usepackage{amssymb}
\usepackage{multirow}
\usepackage{newfloat}
\usepackage{listings}
\DeclareCaptionStyle{ruled}{labelfont=normalfont,labelsep=colon,strut=off} 
\floatstyle{ruled}
\newfloat{listing}{tb}{lst}{}
\floatname{listing}{Listing}

\usepackage{booktabs}

\title{Evaluating Risks in Weak-to-Strong Alignment: A Bias-Variance Perspective}

\title{Evaluating Risks in Weak-to-Strong Alignment: A Bias-Variance Perspective}
\author {
    Hamid Osooli\textsuperscript{\rm 1}\equalcontrib,
    Kareema Batool\textsuperscript{\rm 2},
    Rick Gentry\textsuperscript{\rm 3},
    Tiasa Singha Roy\textsuperscript{\rm 4}\equalcontrib,
    Ashwin Gupta\textsuperscript{\rm 3},
    Anirudha Ramesh\textsuperscript{\rm 3}
}
\affiliations {
    \textsuperscript{\rm 1}University of Illinois Urbana-Champaign\\
    \textsuperscript{\rm 2}Microsoft\\
    \textsuperscript{\rm 3}InstaDeep\\
    \textsuperscript{\rm 4}New York University\\
    hosooli2@illinois.edu, kbatool@microsoft.com, r.gentry@instadeep.com, ts5478@nyu.edu, a.gupta@instadeep.com, a.ramesh@instadeep.com
}

\begin{document}

\maketitle

\begin{abstract}
Weak-to-strong alignment offers a promising route to scalable supervision, but it can fail when a strong model becomes confidently wrong on examples that lie in the weak model's blind spots. Understanding such failures requires going beyond aggregate accuracy, since weak-to-strong errors depend not only on whether the strong model disagrees with the weak model, but also on how confidence and uncertainty are distributed across examples. In this work, we analyze weak-to-strong alignment through a bias--variance--covariance lens that connects misfit theory to practical post-training pipelines. We derive a misfit-based upper bound on weak-to-strong population risk and study its empirical components using continuous confidence scores. We evaluate four weak-to-strong pipelines spanning supervised fine-tuning (SFT), reinforcement learning from human feedback (RLHF), and reinforcement learning from AI feedback (RLAIF) on the PKU-SafeRLHF and HH-RLHF datasets. Using a blind-spot deception metric that isolates cases where the strong model is confidently wrong while the weak model is uncertain, we find that strong-model variance is the quantity most strongly associated with blind-spot deception among the BVC quantities we study. Covariance provides additional but weaker information, indicating that weak--strong dependence matters, but does not by itself explain the observed failures. These results suggest that strong-model variance can serve as an early-warning signal for weak-to-strong deception, while blind-spot evaluation helps distinguish whether failures are inherited from weak supervision or arise in regions of weak-model uncertainty.
\end{abstract}


\section{Introduction}

Weak-to-strong alignment offers a scalable way to train stronger models using supervision from weaker models~\cite{burns2024weak,yangsuper,lyumacpo}. While this approach can reduce reliance on expensive human annotation, it also creates a reliability risk: the strong model may inherit, amplify, or confidently act on errors that the weak model cannot detect. This concern is especially important in preference-based alignment, where the supervision signal is indirect and model confidence can vary substantially across examples.

Prior work has shown that strong models can sometimes outperform weak models when trained on weak labels~\cite{burns2024weak}, and that weak-to-strong failures can appear as deceptive or blind-spot behavior when the weak model is uncertain~\cite{yangsuper}. In parallel, recent theoretical work connects weak-to-strong generalization to misfit and population-risk formulations~\cite{charikar2024quantifying,mulgund2025relating,xu2025emergence}. However, there remains a gap between these theoretical analyses and practical post-training evaluation: existing diagnostics do not directly connect misfit-inspired bias--variance structure with blind-spot deception across both supervised and reinforcement-learning-based alignment pipelines.

Although teacher confidence and transfer under capacity gaps have been studied in knowledge distillation and confidence regularization, our focus is different: we study weak-to-strong alignment as a safety-relevant relabeling problem and use confidence statistics diagnostically rather than as a training regularizer.

In this work, we study weak-to-strong alignment through a confidence-based bias--variance--covariance (BVC) diagnostic framework. We evaluate four weak-to-strong pipelines spanning supervised fine-tuning (SFT), reinforcement learning from human feedback (RLHF), and reinforcement learning from AI feedback (RLAIF) on PKU-SafeRLHF~\cite{dai2024safe} and HH-RLHF~\cite{bai2022training}. We derive a misfit-based upper bound on weak-to-strong population risk and decompose the relevant quantities into bias, variance, and covariance terms. Empirically, we estimate these terms using continuous pairwise confidence scores and compare them against a blind-spot deception metric that isolates cases where the strong model is confidently wrong while the weak model is uncertain.

Our results show that strong-model variance is the quantity most consistently associated with blind-spot deception across the evaluated settings. At the same time, the structure of weak-model uncertainty affects where these failures occur, indicating that deception is not solely a property of strong-model confidence. These findings suggest that confidence-based BVC diagnostics can help identify weak-to-strong failure modes that are hidden by aggregate accuracy alone.

    \textbf{Contributions.}

\textbf{1.} We introduce a blind-spot deception metric that isolates cases where the strong model is confidently wrong while the weak model is uncertain.

\textbf{2.} We develop a confidence-based BVC diagnostic framework that connects misfit-inspired theory with practical weak-to-strong evaluation.

\textbf{3.} We evaluate four weak-to-strong pipelines across SFT, RLHF, and RLAIF settings on two alignment datasets.

\textbf{4.} We find that strong-model variance is the quantity most strongly associated with blind-spot deception among the BVC quantities we study.
\section{Related Work}

\paragraph{Weak-to-strong alignment.}
Weak-to-strong generalization was introduced by \cite{burns2024weak}, who showed that a stronger model can outperform a weaker model when trained on weak labels. Their work focuses primarily on the reward-modeling stage and introduces Performance Gap Recovered (PGR) to measure how much of the strong model's potential performance is recovered under weak supervision. Subsequent work studies weak-to-strong alignment in offline preference-optimization settings. In particular, \cite{yangsuper} examine deception in weak-to-strong training with DPO~\cite{rafailov2023direct} and SimPO~\cite{meng2024simpo}, and propose bootstrapping with intermediate models as a mitigation strategy. In contrast, our work evaluates deception and BVC diagnostics across both supervised and reinforcement-learning-based pipelines, including SFT, RLHF, and RLAIF.

\paragraph{Knowledge distillation, confidence, and capacity gaps.}
Our setting is related to, but distinct from, classical knowledge distillation and confidence-regularization methods. In standard knowledge distillation, a teacher model transfers information to a student through soft targets, often for model compression or improved generalization~\citep{beyer2022knowledge}. Related work has shown that distillation quality depends on teacher consistency, training duration, softened targets, and capacity differences between teacher and student~\citep{beyer2022knowledge,roth2024fantastic}. Other approaches regularize confidence directly: \cite{pereyra2017regularizing} penalize overconfident output distributions, while \cite{yun2020regularizing} use self-knowledge distillation to regularize class-wise predictive distributions and improve calibration. These works show that teacher confidence and softened distributions can significantly affect knowledge transfer. However, weak-to-strong alignment differs from this setting: the weak model is not used primarily to compress knowledge into a smaller student or to provide a general-purpose confidence regularizer. Instead, an aligned but weaker model relabels preference data, and a stronger model is trained on these weak labels in a safety-relevant setting. Our contribution is therefore not a new distillation objective, temperature-scaling rule, or confidence penalty; rather, we provide a diagnostic framework for measuring when weak supervision leads to blind-spot deception in stronger models.

\paragraph{Alternative weak-supervision frameworks.}
Other approaches modify the weak-supervision process itself. \cite{lyumacpo} propose a framework using multiple weak models that provide positive and negative signals, allowing the strong model to contrast between different weak-supervision sources. While this improves alignment quality, it requires generating and evaluating multiple candidate responses per example. From a data-centric perspective, \cite{shinweak} argue that weak-to-strong generalization depends on dataset structure and the overlap between patterns learned by weak and strong models. Their analysis highlights risks such as bias amplification, shortcut learning, and blind spots. Our work is complementary: rather than changing the labeling process or studying dataset structure alone, we analyze how different post-training pipelines affect confidence-based weak-to-strong failure modes.

\paragraph{Misfit and theoretical analyses.}
Several recent works provide theoretical foundations for weak-to-strong generalization. \cite{charikar2024quantifying} relate performance gains to the misfit between weak and strong predictors under squared loss, showing that fine-tuning on weak labels can be interpreted as projecting the weak model onto the strong model's hypothesis class. \cite{mulgund2025relating} extend this misfit-based view beyond squared loss using Bregman divergences. \cite{xu2025emergence} further study weak-to-strong generalization from a bias--variance perspective and connect population-risk gaps to misfit. Our work builds on these theoretical insights by turning bias, variance, and covariance quantities into empirical diagnostics based on model confidence scores, and by relating them to blind-spot deception in practical alignment pipelines.

\section{Preliminaries}

We use a standard pairwise-preference formulation for RLHF-style alignment~\citep{xiong2024iterative}. Let $x\in\mathcal{X}$ denote a prompt and $a=[t_1,t_2,\ldots]\in\mathcal{A}$ denote a generated response consisting of $t_i$ tokens. A preference dataset is defined as tuples $(x,a^1,a^2,y)$, where $a^1$ and $a^2$ are candidate responses and $y$ indicates which response is preferred. We write $a^1\succ a^2$ when $a^1$ is preferred over $a^2$, and denote the underlying human-aligned reward by $r^*(x,a)\in[0,1]$. As in standard RLHF, reward models can be fit from pairwise preferences using the Bradley--Terry model~\citep{bradley1952rank}, and policies can be optimized with a KL-regularized objective; we provide these standard details in supplementary materials for completeness.

\paragraph{Terminology.}
Throughout the paper, we use \emph{weak model} to refer to $\pi^w$, the model or policy used to generate weak labels, and \emph{strong model} to refer to $\pi^s$, the model trained on those weak labels. Earlier terms such as weak teacher or weak supervisor refer
to the same object, but we avoid them for consistency. When discussing classical knowledge distillation in Related Work, we retain the conventional teacher--student terminology only for that separate literature.

    \subsection{Confidence Score}

    For a pair of outputs $a^1 \succ a^2$ associated with prompt $x$, we define the confidence score of the model $\pi$ as:

        \begin{equation}\label{eq:conf-score}
        \mathcal{C}^{\pi}(x) = \sigma(\ell_{\pi}(a^1|x) - \ell_{\pi}(a^2|x)),
    \end{equation}

    \noindent where $\ell^{\pi}(a|x)$ denotes the average log probability of completion $a$ under model~$\pi$:
    
    \begin{equation}\label{eq:avg-log-prob}
        \ell_{\pi}(a|x) = \frac{1}{|a|}\sum_{i=1}^{|a|} log\pi(t_i|x).
    \end{equation}
    
    The sigmoid function $\sigma(\cdot)$ maps the log-probability difference into $(0,1)$, producing a normalized confidence score indicating how strongly $\pi$ prefers the winning completion over the losing one.
    \subsection{Deception Metrics}

We focus on a deception metric tailored to weak-model blind spots. We define \emph{blind-spot deception} as the fraction of strong-model errors for which the strong model is confidently wrong while the weak model remains near the decision boundary. Intuitively, these are cases where the strong model makes a high-confidence mistake precisely in regions where the weak model is uncertain and therefore unable to provide reliable guidance.

Formally, let $\mathcal{C}^{\pi^s}(x)$ and $\mathcal{C}^{\pi^w}(x)$ denote the confidence scores of the strong and weak models on prompt $x$, respectively, where confidence is computed from the pairwise preference score. A blind-spot deceptive case occurs when the strong model is wrong and confidently so, i.e., $\mathcal{C}^{\pi^s}(x) < 0.5 - \tau$, while the weak model lies within an uncertainty band around the decision boundary, i.e., $|\mathcal{C}^{\pi^w}(x)-0.5| < \tau$. We therefore define blind-spot deception at threshold $\tau$ as

\begin{equation}
\label{eq:blindspot_dec}
\resizebox{\columnwidth}{!}{$
d_{\mathrm{BS}}^{\tau}
=
\frac{
\left|\left\{
x :
y^{\pi^s}\neq y^{\mathrm{gt}},\;
\mathcal{C}^{\pi^s}(x)<0.5-\tau,\;
\left|\mathcal{C}^{\pi^w}(x)-0.5\right|<\tau
\right\}\right|
}{
\left|\left\{
x : y^{\pi^s}\neq y^{\mathrm{gt}}
\right\}\right|
}.$
}
\end{equation}

This definition isolates failures that arise in weak-model blind spots, rather than cases where both weak and strong models simply make the same confident mistake. We also consider a broader auxiliary deception metric in the supplementary material.

\section{Problem Formulation}\label{problem_formulation}
\subsection{Population Risk and Misfit}
The human (ground truth) reward function is defined as $r^*(x,a)$. Let $r^w(x,a)$, and $r^s(x,a)$ denote the weak and strong models estimated rewards respectively. For brevity, we omit the explicit dependence on $(x,a)$ and write $r$ instead of $r(x,a)$ when the context is clear. Following~\cite{xu2025emergence}, the \textit{expected population risk} for the policy $\pi$ can be defined as:

    \begin{equation}\label{eq:pop_risk}
        \mathcal{R}(\pi) = \mathbb{E}_{x\sim d_0, a\sim \pi}[L(r^*, \hat{r}^{\pi})],
    \end{equation}

    \noindent where $L(\cdot,\cdot)$ is a \textit{Bregman divergence}. Using squared loss as a special case of Bregman divergence~\cite{xu2025emergence}, the expected population risk for the weak policy $\pi^w$ is defined as:

        \begin{equation}\label{eq:pop_risk_weak}
        \mathcal{R}(\pi^w) = \mathbb{E}_{x\sim d_0, a\sim \pi^w}[(r^* - \hat{r}^{\pi^w})^2],
    \end{equation}

    Similarly we can define the \textit{expected misfit}~\cite{xu2025emergence,mulgund2025relating} or student-supervisor disagreement~\cite{burns2024weak} as:
    
            \begin{equation}
        \mathcal{M}(\pi^s, \pi^w) = \mathbb{E}_{x\sim d_0, a\sim \pi^w}[\big(\hat{r}^{\pi^s} - \hat{r}^{\pi^w}\big)^2].
    \end{equation}
    
    We define weak-to-strong population risk as the off-policy population risk of the strong model under the weak policy’s action distribution. More formally:

    \begin{equation}
    \label{eq:off_policy_risk}
        \mathcal{R}_{w2s}(\pi^s|\pi^w) = \mathbb{E}_{x\sim d_0, a\sim \pi^w}[\big(r^* - \hat{r}^{\pi^s}\big)^2].
    \end{equation}
    
    We establish a misfit-dependent upper bound on the weak-to-strong population risk (please see the proof in the supplementary material). The off-policy population risk of the strong model, evaluated under the weak model’s action distribution, is bounded above by population risk of the weak model with respect to the ground-truth reward plus the expected misfit between the strong and weak reward models.
    
    \begin{equation}
    \label{eq:finalbound}
        \mathcal{R}_{w2s}(\pi^s|\pi^w)\leq (\sqrt{\mathcal{R}(\pi^w)} + \sqrt{\mathcal{M}(\pi^s, \pi^w)})^2
    \end{equation}

\subsection{Bias--Variance--Covariance Decomposition}

For a given prompt $x$, we estimate the conditional reward moments induced by the weak policy $\pi^w(\cdot\mid x)$. Consequently, both the ground truth reward $r^*(x,a)$ and the predicted reward $\hat{r}^{\pi^w}(x,a)$ are random variables. 

We define the expected values (means) over the policy distribution as:
\begin{align}
\bar{r}^* &\triangleq \mathbb{E}_{a\sim\pi^w}[r^*], \\
\bar{r}^{\pi^w} &\triangleq \mathbb{E}_{a\sim\pi^w}[\hat{r}^{\pi^w}].
\end{align}

Using these means, we define the standard squared bias, model variance, target variance, and covariance:
\begin{equation}\label{eq:weak_bias_variance}
    \begin{aligned}
\mathrm{Bias}^2(\hat{r}^{\pi^w})
&\triangleq
(\bar{r}^{\pi^w} - \bar{r}^*)^2, \\
\mathrm{Var}(\hat{r}^{\pi^w})
&\triangleq
\mathbb{E}_{a\sim\pi^w}
\big[
(\hat{r}^{\pi^w} - \bar{r}^{\pi^w})^2
\big], \\
\mathrm{Var}(r^*)
&\triangleq
\mathbb{E}_{a\sim\pi^w}
\big[
(r^* - \bar{r}^*)^2
\big],\\
\mathrm{Cov}(r^*, \hat{r}^{\pi^w})
&\triangleq
\mathbb{E}_{a\sim \pi^w}\big[(r^* - \bar{r}^*)(\hat{r}^{\pi^w} - \bar{r}^{\pi^w})\big].
\end{aligned}
\end{equation}
{These four quantities provide a direct Bias-Variance-Covariance (BVC) interpretation of the weak reward model under the \textit{weak policy distribution}. The squared bias measures the systematic gap between the average predicted reward and the average ground-truth reward, indicating whether the weak reward model consistently over- or under-estimates the true preference signal. The weak model variance measures how much the predicted reward changes across actions sampled from $\pi^w$, capturing instability or sensitivity of the reward estimate. The target variance measures the inherent variability of the true reward across the \textit{same action distribution}, reflecting how difficult the prompt is to evaluate. Finally, the covariance measures the statistical co-variation between the ground-truth reward and the weak reward estimate. This term should not be interpreted in isolation as alignment quality; rather, its effect is meaningful as part of the full BVC decomposition in Eq.~\ref{eq:Rw_bias_var}. For fixed variances and bias, larger positive covariance reduces the squared-risk expression, while low or negative covariance increases it.}
Using these terms we can decompose $\mathcal{R}(\pi^w)$ into bias-variance-covariance terms as (proof in the supplementary material):

\begin{equation}\label{eq:Rw_bias_var}
\resizebox{\columnwidth}{!}{
$    \begin{aligned}
    \mathcal{R}(\pi^w) = & \mathrm{Bias}^2(\hat{r}^{\pi^w})
    + \mathrm{Var}(\hat{r}^{\pi^w}) 
    + \mathrm{Var}(r^*) - 2\mathrm{Cov}(r^*, \hat{r}^{\pi^w}).
    \end{aligned}$
    }
\end{equation}

This decomposition separates average weak-model bias, prediction dispersion under the weak policy, intrinsic target variability, and alignment with the ground-truth signal.

Similarly for $\mathcal{M}$, we define the expected values (means) of the student predictions with respect to the weak policy $\pi^w(\cdot\mid x)$:

\begin{align}
\bar{r}^{\pi^s} &\triangleq \mathbb{E}_{a\sim\pi^w}[\hat{r}^{\pi^s}(x,a)].
\end{align}

We also define relative squared bias, and student variance as:

\begin{equation}\label{eq:strong_bias_variance}
    \begin{aligned}
\mathrm{Bias}^2(\hat{r}^{\pi^s}\!\mid\!\hat{r}^{\pi^w})
&\triangleq
(\bar{r}^{\pi^s} - \bar{r}^{\pi^w})^2, \\
\mathrm{Var}(\hat{r}^{\pi^s})
&\triangleq
\mathbb{E}_{a\sim\pi^w}
\big[
(\hat{r}^{\pi^s}(x,a) - \bar{r}^{\pi^s})^2
\big],\\
\mathrm{Cov}(\hat{r}^{\pi^s}, \hat{r}^{\pi^w})
&\triangleq
\mathbb{E}_{a\sim \pi^w}\big[(\hat{r}^{\pi^s} - \bar{r}^{\pi^s})(\hat{r}^{\pi^w} - \bar{r}^{\pi^w})\big].
\end{aligned}
\end{equation}
{These quantities extend the BVC interpretation from the weak reward model to the weak-to-strong setting by measuring how the student reward estimate $\hat{r}^{\pi^s}$ behaves relative to the weak reward estimate $\hat{r}^{\pi^w}$ over \textit{actions sampled from the weak policy $\pi^w$}. The relative squared bias measures the systematic difference between the average student reward estimate and the average weak reward estimate, capturing whether the student shifts the reward signal away from the weak model. The student variance measures how much the student reward estimate fluctuates across the weak policy’s action distribution, reflecting the stability or sensitivity of the student model. The covariance between $\hat{r}^{\pi^s}$ and $\hat{r}^{\pi^w}$ measures their statistical co-variation around their respective means. We do not interpret this covariance term alone as alignment quality or as a causal explanation of deception. Instead, it is a diagnostic component of the full misfit decomposition in Eq.~\ref{eq:Msw_bias_var}. For fixed bias and variances, larger positive covariance reduces weak--strong disagreement, while lower covariance increases it; however, high covariance may also reflect shared weak-model errors and therefore does not necessarily imply better alignment with the ground-truth reward.}
 This yields the following decomposition of $\mathcal{M}(\pi^s, \pi^w)$ (proof in the supplementary material) into the bias-variance-covariance terms as:

\begin{equation}\label{eq:Msw_bias_var}
\begin{aligned}
    &\mathcal{M}(\pi^s, \pi^w)\\ 
    &= \mathrm{Bias}^2(\hat{r}^{\pi^s} \mid \hat{r}^{\pi^w}) + \mathrm{Var}(\hat{r}^{\pi^s}) + \mathrm{Var}(\hat{r}^{\pi^w})\\ 
    &- 2\mathrm{Cov}(\hat{r}^{\pi^s}, \hat{r}^{\pi^w}).    
\end{aligned}
\end{equation}

This shows that weak–strong disagreement depends not only on average prediction shift, but also on the dispersion of strong and weak predictions and their statistical dependence.

By substituting the bias-variance decomposition into
\ref{eq:finalbound}, we obtain:
    
\begin{equation}
\label{eq:w2sineq}
\resizebox{\columnwidth}{!}{
$\begin{aligned}
&\mathcal{R}_{w2s}(\pi^s \mid \pi^w)\\
&\leq
\Big(\sqrt{\mathrm{Bias}^2(\hat{r}^{\pi^w})
    + \mathrm{Var}(\hat{r}^{\pi^w}) 
    + \mathrm{Var}(r^*) - 2\mathrm{Cov}(r^*, \hat{r}^{\pi^w})}
+\\
&\sqrt{\mathrm{Bias}^2(\hat{r}^{\pi^s} \mid \hat{r}^{\pi^w}) + \mathrm{Var}(\hat{r}^{\pi^s}) + \mathrm{Var}(\hat{r}^{\pi^w}) - 2\mathrm{Cov}(\hat{r}^{\pi^s}, \hat{r}^{\pi^w})}\Big)^2.
\end{aligned}$
}
\end{equation}

\section{Experiments}
    
    \begin{figure*}
        \centering
        \includegraphics[trim=0em 4em 0em 1em, clip, width=0.9\linewidth]{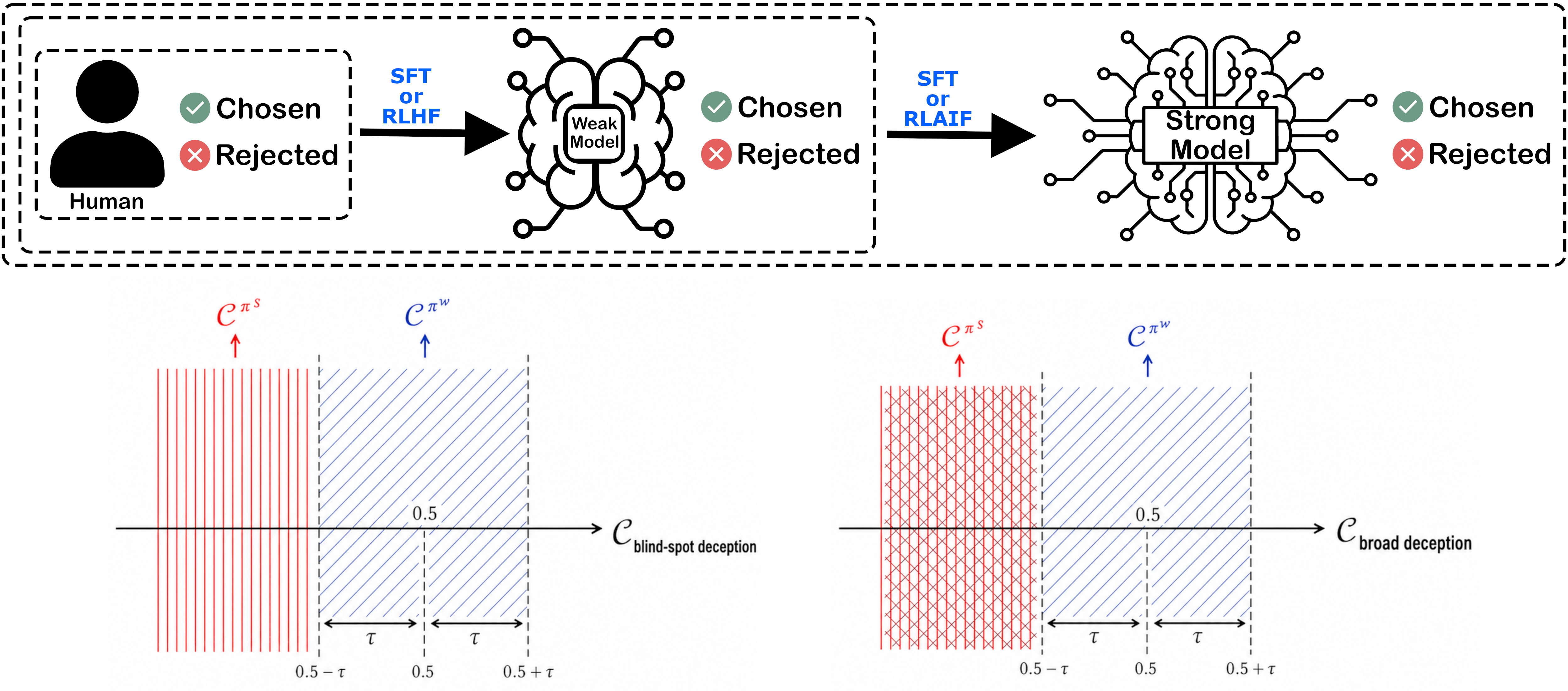}
        \caption{Overview of our weak-to-strong alignment framework and deception criteria. The process begins by training a weak policy $\pi^{w}$ on human-labeled preference data using either SFT or RLHF. The weak policy then provides synthetic supervision for training a strong policy $\pi^{s}$ using either SFT or RLAIF, resulting in four weak-to-strong training variants. The bottom panels illustrate the confidence regions used to define blind-spot deception and broad deception, based on the weak- and strong-policy confidence scores, $\mathcal{C}^{\pi^w}$ and $\mathcal{C}^{\pi^s}$, with decision threshold $0.5$ and tolerance parameter $\tau$.}
        \label{fig:rlaif_w2s}
    \end{figure*}

    Each trained model is assessed to identify where performance drops occur and which steps are most susceptible to learning risks. We investigate four variants of the weak-to-strong alignment pipeline that differ in how the weak and strong models are trained. We partition the dataset $\mathcal{D}=\{(x_i, a_i^1, a_i^2, y_i^{gt})\}_{i=1}^{3N}$ into three disjoint subsets of equal size, $\mathcal{D}_1$, $\mathcal{D}_2$, and $\mathcal{D}_3$. The first subset, $\mathcal{D}_1=\{(x_i, a_i^1, a_i^2, y_i^{gt})\}_{i=1}^N$, contains human-annotated ground-truth preferences and is used to train the initial weak model. $\mathcal{D}_2$ is relabeled using the weak model to generate synthetic weak labels, and $\mathcal{D}_3$ is reserved for held-out evaluation using human annotations. Each configuration can be viewed as a different estimator of the true human-aligned reward function $r^*(x,a)$, with its own bias-variance decomposition. We empirically evaluate each setup through measuring bias, variance, covariance and assigning a deception score.
    
    \textbf{RLHF $\Rightarrow$ RLAIF.} In the first setting, both models are trained through reinforcement learning. We train a reward model on human-labeled preferences $r^*(x,a)$ from $\mathcal{D}_1$ and optimize a weak policy $\pi^{w}(y|x)$ using Proximal Policy Optimization (PPO~\cite{schulman2017proximal}) under the standard Reinforcement Learning from Human Feedback (RLHF~\cite{ouyang2022training}) framework. The resulting weak policy generates synthetic weak labels on $\mathcal{D}_2$ forming $\mathcal{D}_2=\{(x_i, a_i^1, a_i^2, y_i^{w})\}_{i=1}^N$.

    A second reward model is then trained to approximate the weak policy’s preference labels, and a strong policy $\pi^{s}(y|x)$ is optimized on these weak labels using Reinforcement Learning from AI Feedback (RLAIF~\cite{bai2022constitutional}). Both weak and strong policies are evaluated on the held-out set $\mathcal{D}_3=\{(x_i, a_i^1, a_i^2, y_i^{gt})\}_{i=1}^N$, which contains unseen human-labeled examples.

    \textbf{RLHF $\Rightarrow$ SFT.} Here, we replace the reinforcement learning stage of the strong model with supervised fine-tuning (SFT). After training $\pi^{w}(y|x)$ via RLHF on $\mathcal{D}_1$, the weak policy generates synthetic preferences on $\mathcal{D}_2$. The strong model is then fine-tuned on the chosen responses that is, either $a^1$ or $a^2$ where $y^w=1$ without additional reward modeling.
    
    \textbf{SFT $\Rightarrow$ RLAIF.} In this configuration, the weak model is trained by supervised fine-tuning on human-preferred labels from $\mathcal{D}_1$ (i.e., $a^1$ or $a^2$ where $y^{gt}=1$), eliminating the first reward modeling step. The trained weak policy then generates weak labels for $\mathcal{D}_2$, which are used to train a second reward model and an RLAIF-aligned strong policy as in RLHF$\Rightarrow$RLAIF.
    
    \textbf{SFT $\Rightarrow$ SFT.} In our first setup the weak model is fine-tuned on human-preferred labels from $\mathcal{D}_1$, while the strong model is fine-tuned on the weak model’s preferred responses in $\mathcal{D}_2$ (again, $a^1$ or $a^2$ where $y^w=1$). This configuration allows us to isolate the benefits of reinforcement learning and AI-feedback mechanisms introduced in the previous variants.

    \subsection{How SFT Fits the Diagnostic Framework}

A key point in our setup is that the theory is used as a diagnostic framework for policy behavior under pairwise preference evaluation, not as a claim that all pipelines optimize the same reward objective. This matters for supervised fine-tuning (SFT), which does not explicitly optimize a learned reward model in the way RLHF or RLAIF does.

Even so, an SFT-trained policy can still be evaluated on chosen--rejected response pairs through the confidence scores defined earlier. This allows us to compute the same diagnostic quantities, including misfit, bias--variance--covariance terms, and blind-spot deception, for both SFT- and RL-based pipelines. We therefore use a common evaluation framework across all settings while preserving the distinction between how the policies are trained and how they are analyzed.

    \section{Results}\label{sec:results}
\begin{table}[ht]
\centering
\resizebox{\columnwidth}{!}{
\begin{tabular}{c c c c c}
\hline
\textbf{Dataset} & \textbf{W$\boldsymbol{\Rightarrow}$S Pipeline}& \textbf{$\boldsymbol{\mathrm{Var}(\hat{r}^{\pi^s})}$}&\textbf{$\boldsymbol{\mathcal{R}_{w2s}}$}&$\boldsymbol{d_{\mathrm{BS}}^{0.25}}$\\ 
\hline

\textbf{PKU-SafeRLHF}& SFT $\Rightarrow$ SFT
&$0.013787$ & $\underline{0.052639}$ &$\underline{0.002}$\\
& SFT $\Rightarrow$ RLAIF&$0.015994$ & $0.056387$ & $0.019$\\
& RLHF $\Rightarrow$ SFT
&$0.013563$ & $0.075366$ & $0.006$\\
& RLHF $\Rightarrow$ RLAIF& $0.087957$ & $\boldsymbol{0.193993}$ &$\boldsymbol{0.457}$\\
\hline

\textbf{HH-RLHF}& SFT $\Rightarrow$ SFT&$0.025179$ & $\underline{0.085590}$ & $\underline{0.023}$\\
& SFT $\Rightarrow$ RLAIF&$0.040558$ & $0.174069$ & $\boldsymbol{0.162}$\\
& RLHF $\Rightarrow$ SFT&$0.024978$ & $\boldsymbol{0.249482}$ & $0.040$\\
& RLHF $\Rightarrow$ RLAIF&$0.050150$ & $0.170434$ & $0.073$\\
\hline
\end{tabular}}
\caption{Strong-model variance, upper bound, and blind-spot deception at $\tau=0.25$ across training pipelines and datasets. Maximum values per dataset are shown in bold, and minimum values are underlined.}
\label{tab:var-mis-dec}
\end{table}

We organize the results around three questions: which empirical quantities are most strongly associated with blind-spot deception, whether deception can be explained by strong-model confidence dispersion alone, and whether weak-model training changes where blind spots occur. Because this analysis is based on only eight pipeline--dataset settings, we interpret the correlation results as exploratory diagnostic evidence rather than definitive statistical laws. To test whether the main trend is driven by a single extreme point, we also report a filtered robustness analysis that removes the largest PKU-SafeRLHF RLHF$\Rightarrow$RLAIF point and recomputes the correlations over the remaining seven configurations.

\begin{figure*}
    \centering
    \includegraphics[width=0.9\linewidth]{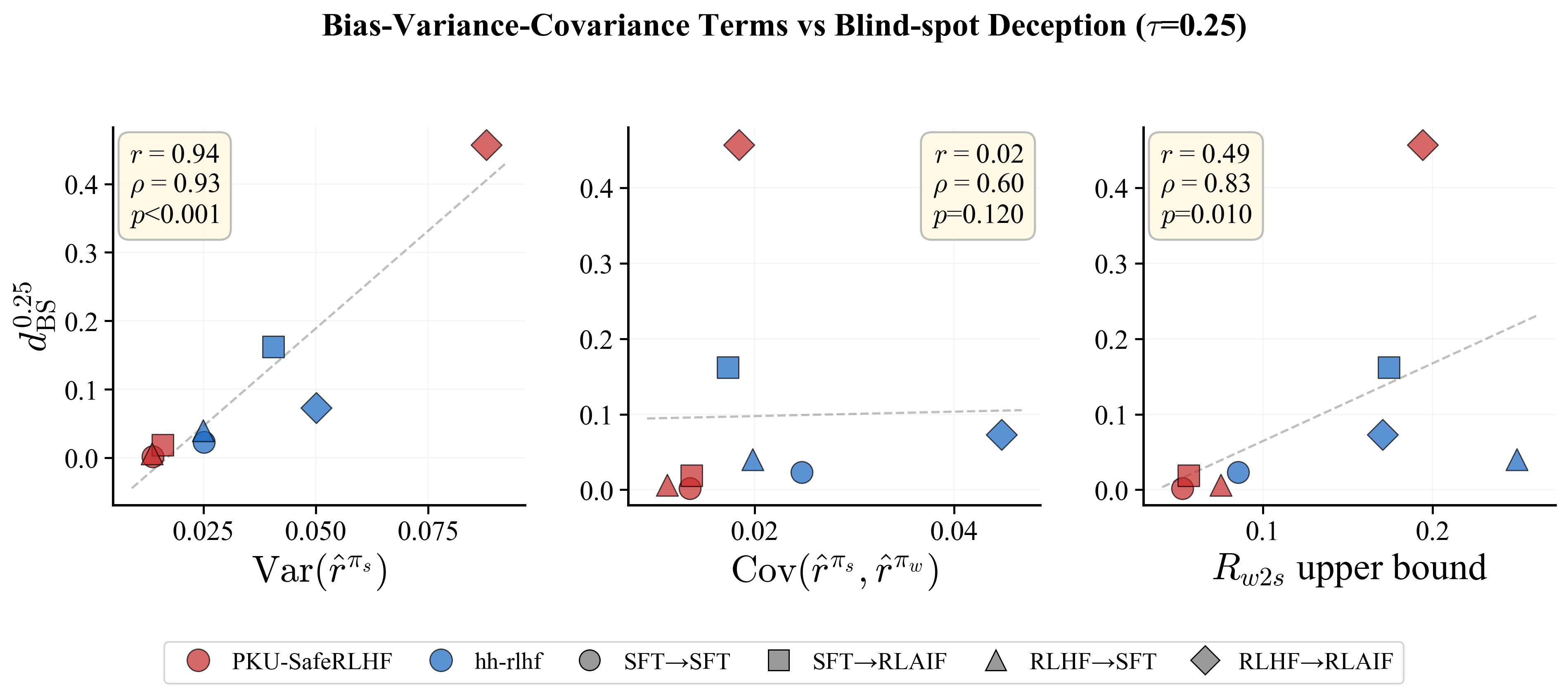}
    \includegraphics[trim=1em 2em 4em 1em, clip, width=0.9\linewidth]{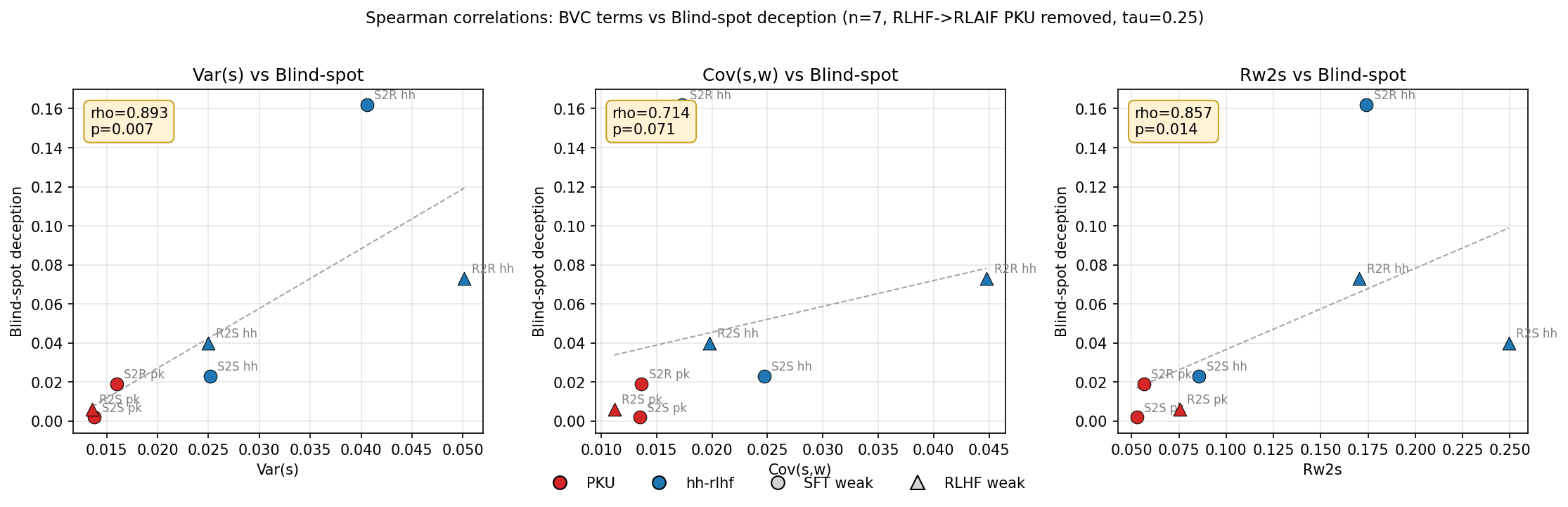}
    \caption{Spearman correlations between blind-spot deception and bias--variance--covariance quantities at $\tau=0.25$. The first analysis uses all eight pipeline/dataset settings, where strong-model variance is more strongly associated with blind-spot deception than weak--strong covariance or the weak-to-strong risk upper bound. The second provides a robustness check after removing the largest PKU-SafeRLHF RLHF$\Rightarrow$RLAIF point; the strong-model variance relationship remains high in the filtered $n=7$ analysis, suggesting that the trend is not solely driven by one extreme configuration.
}
    \label{fig:correlations}
\end{figure*}

\paragraph{Strong-model variance shows the strongest empirical association with blind-spot deception.}
Table~\ref{tab:var-mis-dec} and Fig.~\ref{fig:correlations} show that among the bias--variance--covariance quantities we study, the strongest empirical association with blind-spot deception is observed for the strong-model variance $\mathrm{Var}(\hat{r}^{\pi^s})$. At $\tau=0.25$, its Spearman correlation with $d_{\mathrm{BS}}^{0.25}$ is $\rho=0.929$ ($p=0.001$), compared with $\rho=0.595$ ($p=0.120$) for $\mathrm{Cov}(\hat{r}^{\pi^s},\hat{r}^{\pi^w})$ and $\rho=0.833$ ($p=0.010$) for $\mathcal{R}_{w2s}$.

This pattern is visible at the pipeline level. The most extreme case is RLHF$\Rightarrow$RLAIF on PKU-SafeRLHF, which has the largest strong-model variance ($0.088$) and the largest blind-spot deception score ($0.457$). To check whether the correlation is driven only by this point, we recompute the Spearman correlations after removing it. As shown in Fig.~\ref{fig:correlations}, the strong-model variance correlation remains high and statistically significant under the filtered $n=7$ analysis ($\rho=0.893$, $p=0.007$). This robustness check supports the interpretation that strong-model confidence dispersion is not merely an artifact of one extreme configuration, although the small number of settings means that the result should still be viewed as exploratory.

\paragraph{Deception is not purely variance-driven.}
Although strong-model variance shows the strongest overall association, it is not by itself sufficient to explain all observed behavior. A useful counterexample appears on HH-RLHF. At $\tau=0.25$, SFT$\Rightarrow$RLAIF attains the largest blind-spot deception score on that dataset ($0.162$), even though RLHF$\Rightarrow$RLAIF has both larger strong-model variance ($0.050$ versus $0.040$) and larger weak--strong covariance ($0.045$ versus $0.017$). The same ordering persists across the threshold sweep: on HH-RLHF, SFT$\Rightarrow$RLAIF remains above RLHF$\Rightarrow$RLAIF for blind-spot deception at $\tau \in \{0.10,0.20,0.30,0.40,0.49\}$ (please see the supplementary material). This shows that deception is not solely a function of how spread out the strong model's confidence is. Rather, blind-spot deception also depends on where the weak model is uncertain.

\paragraph{Weak-model training changes blind-spot structure.}
The role of the weak model becomes clearer when comparing the two datasets. On PKU-SafeRLHF, RLHF$\Rightarrow$RLAIF is the dominant blind-spot-deception case across all reported thresholds, reaching $0.477$ at $\tau=0.10$, $0.525$ at $\tau=0.20$, and $0.457$ at $\tau=0.25$. On HH-RLHF, however, the highest blind-spot deception scores arise under SFT$\Rightarrow$RLAIF rather than RLHF$\Rightarrow$RLAIF; for example, at $\tau=0.25$, SFT$\Rightarrow$RLAIF gives $d_{\mathrm{BS}}^{0.25}=0.162$, compared with $0.073$ for RLHF$\Rightarrow$RLAIF.

This reversal is informative because the strong-side training method is fixed in both cases: the strong model is trained with RLAIF, while the weak model is initialized through either RLHF or SFT. Thus, the change in the dominant failure mode cannot be explained only by the strong-side RLAIF stage. Instead, it suggests that the weak model’s training procedure changes the distribution of weak-model confidence and uncertainty. In one dataset, RLHF-based weak supervision creates the blind-spot regions in which the RLAIF-trained strong model becomes confidently wrong; in the other, SFT-based weak supervision creates the more problematic blind spots.

This does not mean that one weak-model training method is universally safer than the other. Rather, the reversal shows that weak-to-strong deception depends on the interaction between dataset structure, weak-model uncertainty, and strong-model confidence. High strong-model variance is most concerning when the corresponding high-confidence errors occur in regions where the weak model is near the decision boundary and therefore cannot reliably distinguish preferred from rejected responses. This also explains why aggregate weak--strong agreement or aggregate risk alone can miss important failure modes: two pipelines may have similar overall performance while placing their errors in different weak-model uncertainty regions.

Because our experiments use only Llama-family models and harmlessness-style preference datasets, we treat this cross-dataset reversal as exploratory evidence. A more complete analysis would require additional model families, datasets, and controlled ablations of the weak-model training stage. Nevertheless, the observed reversal supports our main diagnostic claim: blind-spot deception is not only a property of the strong model, but also depends on how the weak model’s training procedure shapes the regions where weak supervision is uncertain.

\paragraph{Relationship to the weak-to-strong risk bound.}
The misfit-based weak-to-strong risk bound remains informative, but it does not fully determine deception. The $\mathcal{R}_{w2s}$ upper bound is positively correlated with blind-spot deception ($\rho=0.833, p=0.010$), yet it does not recover the cross-dataset reversal described above. For example, RLHF$\Rightarrow$SFT on HH-RLHF has the largest $\mathcal{R}_{w2s}$ upper-bound value ($0.249$), but not the largest blind-spot deception. Conversely, SFT$\Rightarrow$RLAIF on HH-RLHF produces the highest blind-spot deception on that dataset despite a smaller $\mathcal{R}_{w2s}$ upper-bound value ($0.174$). This reinforces the view that blind-spot deception captures a more specific failure mode than aggregate weak-to-strong risk alone.

Overall, the results support two conclusions. First, strong-model variance is the clearest empirical warning signal for blind-spot deception among the decomposition terms we evaluate. Second, weak-model training influences where blind spots arise, so deception cannot be understood as a property of the strong model alone.

\section{Conclusion}

We studied weak-to-strong alignment through a bias--variance--covariance lens that connects misfit-based theory to practical post-training pipelines. Across four weak-to-strong settings spanning SFT, RLHF, and RLAIF, our results show that blind-spot deception is more strongly associated with strong-model variance than with covariance or aggregate weak-to-strong risk alone. At the same time, the cross-dataset reversals indicate that deception is shaped not only by strong-model confidence, but also by where weak-model uncertainty creates blind spots.

These findings suggest two practical lessons. First, strong-model variance may serve as a useful candidate early-warning signal, since highly dispersed confidence after training is associated with elevated blind-spot deception in our experiments. Second, the blind-spot deception metric offers a practical diagnostic for evaluating weak--strong model pairs without requiring a ceiling model, using only confidence scores on a held-out preference set. However, our results stop at diagnosis: we do not yet demonstrate that directly optimizing, regularizing, or selecting models based on this signal reduces deception.

More broadly, our results suggest that evaluating weak-to-strong alignment requires going beyond aggregate transfer success or weak--strong agreement. A strong model may appear to generalize beyond the weak model while still failing systematically in the weak model's blind spots. Future work should test whether strong-model variance can be used in closed-loop mitigation strategies, such as early stopping, checkpoint selection, confidence regularization, or uncertainty-aware data selection, and whether these patterns extend beyond Llama-family models and harmlessness-focused preference datasets.

\bibliography{aaai2027}

\clearpage
\appendix
\section*{Supplementary Material}

\section{Proof of the Misfit-Based Risk Bound}
Starting from the definitions of $\pi^w$, $\mathcal{M}(\pi^s, \pi^w)$, and $\mathcal{R}_{w2s}(\pi^s|\pi^w)$ we derive the bound as follows.

By adding and subtracting $\hat{r}^{\pi^s}$ we obtain:

    \begin{equation}\label{eq:pop_risk_strong_mid}
        \begin{aligned}
            &\mathcal{R}(\pi^w)\\ 
            &= \mathbb{E}_{x\sim d_0, a\sim \pi^w}[\big(r^* - \hat{r}^{\pi^w}\big)^2]\\
            &= \mathbb{E}_{x\sim d_0, a\sim \pi^w}[\big(r^* - \hat{r}^{\pi^s} + \hat{r}^{\pi^s} - \hat{r}^{\pi^w}\big)^2]\\
            &= \mathbb{E}_{x\sim d_0, a\sim \pi^w}[\big(r^* - \hat{r}^{\pi^s}\big)^2] + \mathbb{E}_{x\sim d_0, a\sim \pi^w}[\big(\hat{r}^{\pi^s} - \hat{r}^{\pi^w}\big)^2]\\ &+ 
            2\mathbb{E}_{x\sim d_0, a\sim \pi^w}[\big(r^* - \hat{r}^{\pi^s}\big)\big(\hat{r}^{\pi^s} - \hat{r}^{\pi^w}\big)],
        \end{aligned}
    \end{equation}

    Let 
    \begin{equation}
        \epsilon = \mathbb{E}_{x\sim d_0, a\sim \pi^w}[\big(r^* - \hat{r}^{\pi^s}\big)\big(\hat{r}^{\pi^s} - \hat{r}^{\pi^w}\big)].
    \end{equation} 
    Then:

    \begin{equation}
    \mathcal{R}_{w2s}(\pi^s|\pi^w) = \mathcal{R}(\pi^w) - \mathcal{M}(\pi^s, \pi^w)-2\epsilon.
    \end{equation}
    
    By the Cauchy-Schwarz inequality,
    
    \begin{equation}
    \begin{aligned}
        |\mathbb{E}_{x\sim d_0, a\sim \pi^w}[\big(r^* - \hat{r}^{\pi^s}\big)\big(\hat{r}^{\pi^s} - \hat{r}^{\pi^w}\big)]|\leq\\\sqrt{\mathbb{E}_{x\sim d_0, a\sim \pi^w}[\big(r^* - \hat{r}^{\pi^s}\big)^2]\mathbb{E}_{x\sim d_0, a\sim \pi^w}[\big(\hat{r}^{\pi^s} - \hat{r}^{\pi^w}\big)^2]}
    \end{aligned}
    \end{equation}
    
    Multiplying both sides to 2 we have:

        \begin{equation}
        \label{eq:cauchy-schwarz}
    2|\epsilon|\leq2\sqrt{\mathcal{R}_{w2s}(\pi^s|\pi^w)\mathcal{M}(\pi^s, \pi^w)}.
    \end{equation}
    
    Therefore from \ref{eq:pop_risk_strong_mid} and \ref{eq:cauchy-schwarz} we can say:

    \begin{equation}\label{eq:w2s_pop_risk_ineq}
    \resizebox{\columnwidth}{!}{
        $\mathcal{R}_{w2s}(\pi^s|\pi^w) \leq \mathcal{R}(\pi^w) - \mathcal{M}(\pi^s, \pi^w)+2\sqrt{\mathcal{R}_{w2s}(\pi^s|\pi^w)\mathcal{M}(\pi^s, \pi^w)}.$
        }
    \end{equation}

    Reordering the terms we have:

    \begin{equation}\label{eq:w2s_pop_risk_ineq_reordered}
    \resizebox{\columnwidth}{!}{
$    \begin{aligned}
        \mathcal{R}_{w2s}(\pi^s|\pi^w) + \mathcal{M}(\pi^s, \pi^w) - 2\sqrt{\mathcal{R}_{w2s}(\pi^s|\pi^w)\mathcal{M}(\pi^s, \pi^w)}&\leq \mathcal{R}(\pi^w),\\
        (\sqrt{\mathcal{R}_{w2s}(\pi^s|\pi^w)} - \sqrt{\mathcal{M}(\pi^s, \pi^w)})^2&\leq \mathcal{R}(\pi^w),
    \end{aligned}$
    }
    \end{equation}
    Taking square roots from both sides we have:

    \begin{equation}
    \label{eq:finalbound_appendix}
    \begin{aligned}
        |\sqrt{\mathcal{R}_{w2s}(\pi^s|\pi^w)} - \sqrt{\mathcal{M}(\pi^s, \pi^w)}|\leq \sqrt{\mathcal{R}(\pi^w)}\\
        \sqrt{\mathcal{R}_{w2s}(\pi^s|\pi^w)}\leq \sqrt{\mathcal{R}(\pi^w)} + \sqrt{\mathcal{M}(\pi^s, \pi^w)}\\
        \mathcal{R}_{w2s}(\pi^s|\pi^w)\leq (\sqrt{\mathcal{R}(\pi^w)} + \sqrt{\mathcal{M}(\pi^s, \pi^w)})^2
    \end{aligned}
    \end{equation}

\section{Proof of the Bias–Variance–Covariance Decompositions}

The risk $\mathcal{R}(\pi^w)$ decomposes as follows:

\begin{equation}\label{eq:Rw_bias_var_decomp}
    \begin{aligned}
    &\mathcal{R}(\pi^w)\\ 
    &= \mathbb{E}_{x\sim d_0}\mathbb{E}_{a\sim \pi^w}\big[\big(r^* - \hat{r}^{\pi^w}\big)^2\big] \\
    &= \mathbb{E}_{x\sim d_0}\mathbb{E}_{a\sim \pi^w}\big[\big( (r^* - \bar{r}^*) - (\hat{r}^{\pi^w} - \bar{r}^{\pi^w}) + (\bar{r}^* - \bar{r}^{\pi^w}) \big)^2\big] \\
    &=\underbrace{\mathbb{E}_{x\sim d_0}\mathbb{E}_{a\sim \pi^w}\big[(r^* - \bar{r}^*)^2\big]}_{\text{Target Variance}} + \underbrace{\mathbb{E}_{x\sim d_0}\mathbb{E}_{a\sim \pi^w}\big[(\hat{r}^{\pi^w} - \bar{r}^{\pi^w})^2\big]}_{\text{Teacher Variance}}\\
    &+ \underbrace{\mathbb{E}_{x\sim d_0}\mathbb{E}_{a\sim \pi^w}\big[(\bar{r}^* - \bar{r}^{\pi^w})^2\big]}_{\text{Squared Bias}}\\
&+ 2\mathbb{E}_{x\sim d_0}\mathbb{E}_{a\sim \pi^w}\big[(r^* - \bar{r}^*)(\bar{r}^* - \bar{r}^{\pi^w})\big] \\
    &- 2\mathbb{E}_{x\sim d_0}\mathbb{E}_{a\sim \pi^w}\big[(\hat{r}^{\pi^w} - \bar{r}^{\pi^w})(\bar{r}^* - \bar{r}^{\pi^w})\big] \\
    &- 2\mathbb{E}_{x\sim d_0}\mathbb{E}_{a\sim \pi^w}\big[(r^* - \bar{r}^*)(\hat{r}^{\pi^w} - \bar{r}^{\pi^w})\big].
\end{aligned}
\end{equation}

\noindent In \ref{eq:Rw_bias_var_decomp}, the terms involving the constant differences $(\bar{r}^* - \bar{r}^{\pi^w})$ vanish due to the linearity of expectation. Specifically, for any random variable $Z$ and its mean $\bar{Z}$, $\mathbb{E}[Z - \bar{Z}] = 0$. Therefore:
$$
\mathbb{E}_{a\sim \pi^w}[r^* - \bar{r}^*] = 0 \quad \text{and} \quad \mathbb{E}_{a\sim \pi^w}[\hat{r}^{\pi^w} - \bar{r}^{\pi^w}] = 0.
$$
The remaining cross term is, by definition, the covariance between the ground truth and the predictor scaled by $-2$.

Thus, we arrive at the final decomposition:

\begin{equation}\label{eq:Rw_bias_var_appendix}
    \begin{aligned}
    \mathcal{R}(\pi^w) = & \mathrm{Bias}^2(\hat{r}^{\pi^w})
    + \mathrm{Var}(\hat{r}^{\pi^w}) 
    + \mathrm{Var}(r^*) - 2\mathrm{Cov}(r^*, \hat{r}^{\pi^w}).
    \end{aligned}
\end{equation}

To decompose the term inside the expectation in $\mathcal{M}(\pi^s, \pi^w)$, we add and subtract these means:

\begin{equation}\label{eq:Msw_full_derivation}
    \begin{aligned}
    &\mathcal{M}(\pi^s, \pi^w)\\ 
    &= \mathbb{E}_{x\sim d_0, a\sim \pi^w}[\big(\hat{r}^{\pi^s}(x,a) - \hat{r}^{\pi^w}\big)^2]\\
    &= \mathbb{E}_{x\sim d_0}\mathbb{E}_{a\sim \pi^w}\big[ \big( (\hat{r}^{\pi^s} - \bar{r}^{\pi^s}) - (\hat{r}^{\pi^w} - \bar{r}^{\pi^w}) + (\bar{r}^{\pi^s} - \bar{r}^{\pi^w}) \big)^2 \big] \\
    &= \underbrace{\mathbb{E}_{x\sim d_0}[(\bar{r}^{\pi^s} - \bar{r}^{\pi^w})^2]}_{\text{Squared Bias}} 
    + \underbrace{\mathbb{E}_{x\sim d_0}\mathbb{E}_{a\sim \pi^w}[(\hat{r}^{\pi^s} - \bar{r}^{\pi^s})^2]}_{\text{Student Variance}} 
    + \\
    &\underbrace{\mathbb{E}_{x\sim d_0}\mathbb{E}_{a\sim \pi^w}[(\hat{r}^{\pi^w} - \bar{r}^{\pi^w})^2]}_{\text{Teacher Variance}} \\
    &+ 2\mathbb{E}_{x\sim d_0}\mathbb{E}_{a\sim \pi^w}\big[(\hat{r}^{\pi^s} - \bar{r}^{\pi^s})(\bar{r}^{\pi^s} - \bar{r}^{\pi^w})\big] \\
    &- 2\mathbb{E}_{x\sim d_0}\mathbb{E}_{a\sim \pi^w}\big[(\hat{r}^{\pi^s} - \bar{r}^{\pi^s})(\hat{r}^{\pi^w} - \bar{r}^{\pi^w})\big] \\
    &- 2\mathbb{E}_{x\sim d_0}\mathbb{E}_{a\sim \pi^w}\big[(\hat{r}^{\pi^w} - \bar{r}^{\pi^w})(\bar{r}^{\pi^s} - \bar{r}^{\pi^w})\big].
    \end{aligned}
\end{equation}

The cross terms involving the constant mean difference $(\bar{r}^{\pi^s} - \bar{r}^{\pi^w})$ vanish due to the linearity of expectation. For example:
$$
2(\bar{r}^{\pi^s} - \bar{r}^{\pi^w}) \cdot \mathbb{E}_{a\sim\pi^w}[\hat{r}^{\pi^s}(x,a) - \bar{r}^{\pi^s}] = 2(\bar{r}^{\pi^s} - \bar{r}^{\pi^w}) \cdot 0 = 0.
$$
However, the cross term between the student and teacher variations does not vanish, as they are correlated through the shared action $a$:
\begin{equation}
    \begin{aligned}
    &-2\mathbb{E}_{a\sim\pi^w}\big[(\hat{r}^{\pi^s}(x,a) - \bar{r}^{\pi^s})(\hat{r}^{\pi^w}(x,a) - \bar{r}^{\pi^w})\big]\\
    &= -2\mathrm{Cov}_{a\sim\pi^w}(\hat{r}^{\pi^s}, \hat{r}^{\pi^w}).
\end{aligned}
\end{equation}

Substituting these back into equation~\ref{eq:Msw_full_derivation}, we obtain:
\begin{equation}\label{eq:Msw_bias_var_appendix}
\begin{aligned}
        &\mathcal{M}(\pi^s, \pi^w)\\ 
        &= \mathrm{Bias}^2(\hat{r}^{\pi^s} \mid \hat{r}^{\pi^w}) + \mathrm{Var}(\hat{r}^{\pi^s}) + \mathrm{Var}(\hat{r}^{\pi^w})\\ &- 2\mathrm{Cov}(\hat{r}^{\pi^s}, \hat{r}^{\pi^w}).
\end{aligned}
\end{equation}

\section{Broad Deception}

For completeness, we also consider a broader variant of the deception, which we refer to as \emph{broad deception}. This metric includes all strong-model errors in which the strong model is confidently wrong and the weak model is not confidently correct. In contrast to blind-spot deception, this broader formulation may also include cases where the weak model is itself confidently wrong, and therefore captures both blind-spot failures and inherited weak-model errors. We define broad deception as

\begin{equation}
\label{eq:broad_dec}
\begin{aligned}
    &d_{\mathrm{Broad}}^{\tau}\\
&=
\frac{
\left|\left\{
x :
y^{\pi^s}\neq y^{\mathrm{gt}},\;
\mathcal{C}^{\pi^s}(x)<0.5-\tau,\;
\mathcal{C}^{\pi^w}(x)<0.5+\tau
\right\}\right|
}{
\left|\left\{
x : y^{\pi^s}\neq y^{\mathrm{gt}}
\right\}\right|
}.
\end{aligned}
\end{equation}

We treat blind-spot deception as the primary metric, since it more directly captures the failure mode of interest in weak-to-strong supervision, as confident strong-model errors that occur when the weak teacher is uncertain.

\section{Additional Results}

The results in this supplementary material provide robustness checks and supporting detail for the main findings including the broader deception metric, the full bias--variance--covariance decomposition, and sensitivity to the confidence threshold $\tau$.

\subsection{Broad-Deception Correlations}

\begin{figure*}[h]
    \centering
    \includegraphics[width=0.9\linewidth]{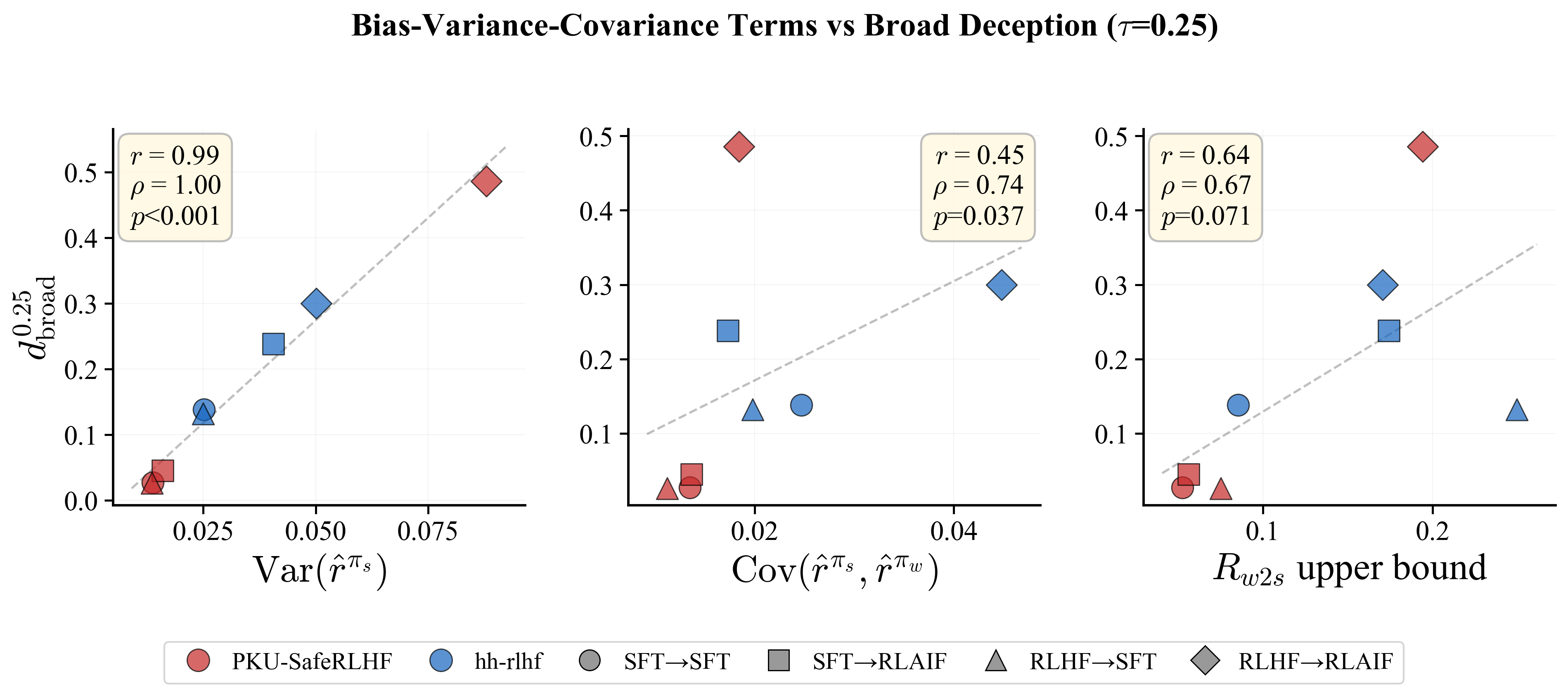}
    \caption{Spearman correlations between broad deception and selected bias--variance--covariance quantities across the eight pipeline/dataset settings at $\tau=0.25$. Among the quantities evaluated, strong-model variance shows the strongest association with broad deception.}
    \label{fig:correlations_extra}
\end{figure*}

Figure~\ref{fig:correlations_extra} reports the Spearman correlations between broad deception and the selected BVC quantities across the eight dataset--pipeline settings at $\tau=0.25$. Strong-model variance exhibits the strongest association with broad deception ($\rho=1.000$, $p<0.001$), followed by weak--strong covariance ($\rho=0.738$, $p=0.037$). The weak-to-strong risk bound also shows a positive association ($\rho=0.667$), although it is not statistically significant at the conventional $0.05$ level ($p=0.071$). These results are consistent with the blind-spot analysis in showing that dispersion in strong-model confidence is closely associated with deceptive errors, while covariance and overall risk provide additional but weaker signals. Given the small number of pipeline--dataset settings, we interpret these correlations as descriptive associations rather than evidence of a causal relationship.

\subsection{Full Bias--Variance--Covariance Decomposition}

Table~\ref{tab:bias-var-cov} reports the full confidence-based BVC decomposition together with the resulting weak-to-strong risk bound and both deception metrics. The components vary substantially across training pipelines and datasets, indicating that the observed failures cannot be explained by bias or covariance alone. The clearest case occurs for RLHF$\Rightarrow$RLAIF on PKU-SafeRLHF, which has substantially higher strong-model variance ($0.087957$) than the other pipelines and simultaneously exhibits the highest weak-to-strong risk bound, blind-spot deception, and broad deception on that dataset. On HH-RLHF, however, the pattern is more distributed across the decomposition terms: SFT$\Rightarrow$RLAIF has the highest blind-spot deception, RLHF$\Rightarrow$RLAIF has the highest broad deception, and RLHF$\Rightarrow$SFT has the largest risk bound. Overall, the decomposition highlights that weak-to-strong failure modes depend on the interaction among prediction dispersion, weak--strong dependence, and the structure of weak-model uncertainty rather than on any single BVC component in isolation.

\begin{table*}[h]
\centering
\resizebox{\textwidth}{!}{
\begin{tabular}{c c c c c c c c}
\hline
&&\textbf{$\boldsymbol{\mathrm{Bias}^2(\hat{r}^{\pi^w})}$}&\textbf{$\boldsymbol{\mathrm{Var}(\hat{r}^{\pi^w})}$}& \textbf{$\boldsymbol{\mathrm{Cov}(r^*, \hat{r}^{\pi^w})}$}&&&\\ 
\textbf{Dataset} & \textbf{W$\boldsymbol{\Rightarrow}$S Pipeline}& \textbf{$\boldsymbol{\mathrm{Bias}^2(\hat{r}^{\pi^s} \mid \hat{r}^{\pi^w})}$}&\textbf{$\boldsymbol{\mathrm{Var}(\hat{r}^{\pi^s})}$}& \textbf{$\boldsymbol{\mathrm{Cov}(\hat{r}^{\pi^s}, \hat{r}^{\pi^w})}$}&\textbf{$\boldsymbol{\mathcal{R}_{w2s}}$}&$\boldsymbol{d_{\mathrm{BS}}^{0.25}}$&$\boldsymbol{d_{\mathrm{Broad}}^{0.25}}$\\ 
& & \textbf{$\boldsymbol{--}$}&\textbf{$\boldsymbol{\mathrm{Var}(r^*)}$}& \textbf{$\boldsymbol{--}$}&&&\\
\hline

\multirow{12}{*}{\rotatebox[origin=c]{90}{\textbf{PKU-SafeRLHF}}}

& SFT $\Rightarrow$ SFT
& $2.0\times10^{-6}$ & $0.015275$ & $0.000071$ &\\
&
& $9.49\times10^{-4}$ & $0.013787$ & $0.013489$ & $\underline{0.052639}$ & $\underline{0.002}$ & $0.027$\\
&
& -- & $0.015268$ & -- &\\

& SFT $\Rightarrow$ RLAIF
& $2.0\times10^{-6}$ & $0.015275$ & $0.000071$ &\\
&
& $1.0\times10^{-6}$ & $0.015994$ & $0.013644$ & $0.056387$ & $0.019$ & $0.045$\\
&
& -- & $0.015268$ & -- &\\

& RLHF $\Rightarrow$ SFT
& $4.9\times10^{-5}$ & $0.016520$ & $0.000080$ &\\
&
& $9.82\times10^{-4}$ & $0.013563$ & $0.011184$ & $0.075366$ & $0.006$ & $\underline{0.026}$\\
&
& -- & $0.016454$ & -- &\\

& RLHF $\Rightarrow$ RLAIF
& $4.5\times10^{-5}$ & $0.016399$ & $0.000096$ &\\
&
& $2.2\times10^{-5}$ & $0.087957$ & $0.018408$ & $\boldsymbol{0.193993}$ & $\boldsymbol{0.457}$ & $\boldsymbol{0.486}$\\
&
& -- & $0.016337$ & -- &\\

\hline

\multirow{12}{*}{\rotatebox[origin=c]{90}{\textbf{HH-RLHF}}}

& SFT $\Rightarrow$ SFT
& $1.6\times10^{-5}$ & $0.027040$ & $-0.000693$ &\\
&
& $4.27\times10^{-4}$ & $0.025179$ & $0.024694$ & $\underline{0.085590}$ & $\underline{0.023}$ & $0.138$\\
&
& -- & $0.027001$ & -- &\\

& SFT $\Rightarrow$ RLAIF
& $1.6\times10^{-5}$ & $0.027040$ & $-0.000693$ &\\
&
& $4.3\times10^{-5}$ & $0.040558$ & $0.017304$ & $0.174069$ & $\boldsymbol{0.162}$ & $0.238$\\
&
& -- & $0.027001$ & -- &\\

& RLHF $\Rightarrow$ SFT
& $7.1\times10^{-5}$ & $0.050055$ & $0.002054$ &\\
&
& $5.04\times10^{-4}$ & $0.024978$ & $0.019778$ & $\boldsymbol{0.249482}$ & $0.040$ & $\underline{0.132}$\\
&
& -- & $0.049957$ & -- &\\

& RLHF $\Rightarrow$ RLAIF
& $7.1\times10^{-5}$ & $0.050055$ & $0.002054$ &\\
&
& $1.0\times10^{-6}$ & $0.050150$ & $0.044794$ & $0.170434$ & $0.073$ & $\boldsymbol{0.300}$\\
&
& -- & $0.049957$ & -- &\\

\hline
\end{tabular}}
\caption{Bias-variance-covariance decomposition of weak and strong models, the resulting weak-to-strong alignment upper bound $\mathcal{R}_{w2s}$, with blind-spot ($d_{\mathrm{BS}}$) and broad ($d_{\mathrm{Broad}}$) deception scores at $\tau=0.25$ across training pipelines and datasets. Maximum values per dataset are shown in bold, and minimum values are underlined.}
\label{tab:bias-var-cov}
\end{table*}

\subsection{Threshold Sensitivity of Blind-Spot Deception}

Table~\ref{tab:deception-multitau} evaluates the sensitivity of blind-spot deception to the confidence threshold $\tau$. The main cross-pipeline pattern remains stable across the evaluated thresholds: RLHF$\Rightarrow$RLAIF consistently produces the highest blind-spot deception on PKU-SafeRLHF, whereas SFT$\Rightarrow$RLAIF is consistently highest on HH-RLHF. The absolute scores generally become small at stricter thresholds, reflecting the decreasing number of strong-model errors that satisfy the high-confidence error condition. On PKU-SafeRLHF, RLHF$\Rightarrow$RLAIF remains a clear outlier, reaching $0.525$ at $\tau=0.20$ before decreasing to $0.023$ at $\tau=0.49$. These results indicate that the relative vulnerability of the pipelines is not an artifact of selecting a single value of $\tau$, and further support the observation that the training procedure used for the weak model affects the structure of blind-spot failures.

\begin{table*}[h]
\centering
\resizebox{\textwidth}{!}{
\begin{tabular}{c c c c c c c}
\hline
\textbf{Dataset} & \textbf{Pipeline (denominator)} 
& $\boldsymbol{\tau=0.10}$ 
& $\boldsymbol{\tau=0.20}$ 
& $\boldsymbol{\tau=0.30}$ 
& $\boldsymbol{\tau=0.40}$ 
& $\boldsymbol{\tau=0.49}$ \\
\hline

\multirow{4}{*}{\textbf{PKU-SafeRLHF}}\\
\multirow{4}{*}{\textbf{(score)}}
& SFT $\Rightarrow$ SFT 
& 0.047 & 0.016 & 0.000 & 0.000 & 0.000 \\

& SFT $\Rightarrow$ RLAIF 
& 0.124 & 0.049 & 0.009 & 0.001 & 0.000 \\

& RLHF $\Rightarrow$ SFT 
& 0.122 & 0.031 & 0.002 & 0.000 & 0.000 \\

& RLHF $\Rightarrow$ RLAIF 
& \textbf{0.477} & \textbf{0.525} & \textbf{0.362} & \textbf{0.189} & \textbf{0.023} \\
\hline
\multirow{4}{*}{\textbf{PKU-SafeRLHF}}\\
\multirow{4}{*}{\textbf{(numerator)}}
& SFT $\Rightarrow$ SFT (2494) 
& 118 & 40 & 1 & 1 & 0 \\

& SFT $\Rightarrow$ RLAIF (2513) 
& 311 & 123 & 22 & 2 & 0 \\

& RLHF $\Rightarrow$ SFT (2498) 
& 304 & 77 & 5 & 0 & 0 \\

& RLHF $\Rightarrow$ RLAIF (2339) 
& \textbf{1115} & \textbf{1229} & \textbf{847} & \textbf{442} & \textbf{53} \\

\hline
\multirow{4}{*}{\textbf{HH-RLHF}}\\
\multirow{4}{*}{\textbf{(score)}}
& SFT $\Rightarrow$ SFT 
& 0.065 & 0.031 & 0.016 & 0.003 & 0.000 \\

& SFT $\Rightarrow$ RLAIF 
& \textbf{0.261} & \textbf{0.219} & \textbf{0.127} & \textbf{0.051} & \textbf{0.009} \\

& RLHF $\Rightarrow$ SFT 
& 0.110 & 0.059 & 0.020 & 0.002 & 0.000 \\

& RLHF $\Rightarrow$ RLAIF 
& 0.133 & 0.095 & 0.061 & 0.027 & 0.006 \\
\hline
\multirow{4}{*}{\textbf{HH-RLHF}}\\
\multirow{4}{*}{\textbf{(numerator)}}
& SFT $\Rightarrow$ SFT (2491) 
& 162 & 78 & 40 & 7 & 0 \\

& SFT $\Rightarrow$ RLAIF (2802) 
& \textbf{731} & \textbf{615} & \textbf{355} & \textbf{142} & \textbf{25} \\

& RLHF $\Rightarrow$ SFT (2500) 
& 276 & 148 & 49 & 4 & 0 \\

& RLHF $\Rightarrow$ RLAIF (2722)
& 361 & 258 & 166 & 73 & 15 \\

\hline
\end{tabular}}
\caption{Blind-spot deception scores and counts across pipelines for different confidence thresholds $\tau$. The qualitative ranking is stable across thresholds: RLHF $\Rightarrow$ RLAIF is consistently highest on PKU-SafeRLHF, whereas SFT $\Rightarrow$ RLAIF is consistently highest on HH-RLHF. This supports the claim that weak-model training affects where blind spots arise.}
\label{tab:deception-multitau}
\end{table*}

\subsection{Threshold Sensitivity of Broad Deception}

Table~\ref{tab:broad-multitau} reports the corresponding threshold sensitivity for broad deception. Broad-deception scores decrease substantially as $\tau$ increases, showing that the absolute magnitude of this metric is sensitive to the confidence threshold. Nevertheless, the relative ordering is highly consistent: RLHF$\Rightarrow$RLAIF has the highest broad-deception score across all evaluated thresholds on both datasets. For example, its score decreases from $0.736$ to $0.027$ on PKU-SafeRLHF and from $0.663$ to $0.027$ on HH-RLHF as $\tau$ increases from $0.10$ to $0.49$. Notably, this differs from the blind-spot metric on HH-RLHF, where SFT$\Rightarrow$RLAIF is consistently highest. This distinction reflects the broader definition of $d_{\mathrm{Broad}}$, which can include confident strong-model errors associated with weak-model errors in addition to failures occurring specifically in regions of weak-model uncertainty.

\begin{table*}[h]
\centering
\resizebox{\textwidth}{!}{
\begin{tabular}{c c c c c c c}
\hline
\textbf{Dataset} & \textbf{Pipeline (denominator)} 
& $\boldsymbol{\tau=0.10}$ 
& $\boldsymbol{\tau=0.20}$ 
& $\boldsymbol{\tau=0.30}$ 
& $\boldsymbol{\tau=0.40}$ 
& $\boldsymbol{\tau=0.49}$ \\
\hline

\multirow{4}{*}{\textbf{PKU-SafeRLHF}}\\
\multirow{4}{*}{\textbf{(score)}}
& SFT $\Rightarrow$ SFT 
& 0.370 & 0.072 & 0.012 & 0.003 & 0.000 \\

& SFT $\Rightarrow$ RLAIF 
& 0.413 & 0.104 & 0.023 & 0.003 & 0.000 \\

& RLHF $\Rightarrow$ SFT 
& 0.361 & 0.070 & 0.012 & 0.002 & 0.000 \\

& RLHF $\Rightarrow$ RLAIF 
& \textbf{0.736} & \textbf{0.582} & \textbf{0.384} & \textbf{0.198} & \textbf{0.027} \\

\hline
\multirow{4}{*}{\textbf{PKU-SafeRLHF}}\\
\multirow{4}{*}{\textbf{(numerator)}}
& SFT $\Rightarrow$ SFT (2494)
& 922 & 180 & 30 & 7 & 0 \\

& SFT $\Rightarrow$ RLAIF (2513)
& 1038 & 262 & 57 & 8 & 0 \\

& RLHF $\Rightarrow$ SFT (2498)
& 901 & 174 & 29 & 6 & 0 \\

& RLHF $\Rightarrow$ RLAIF (2339)
& \textbf{1722} & \textbf{1361} & \textbf{898} & \textbf{462} & \textbf{62} \\

\hline
\multirow{4}{*}{\textbf{HH-RLHF}}\\
\multirow{4}{*}{\textbf{(score)}}
& SFT $\Rightarrow$ SFT 
& 0.492 & 0.219 & 0.084 & 0.017 & 0.003 \\

& SFT $\Rightarrow$ RLAIF 
& 0.553 & 0.338 & 0.172 & 0.064 & 0.012 \\

& RLHF $\Rightarrow$ SFT 
& 0.445 & 0.205 & 0.077 & 0.018 & 0.003 \\

& RLHF $\Rightarrow$ RLAIF 
& \textbf{0.663} & \textbf{0.398} & \textbf{0.228} & \textbf{0.109} & \textbf{0.027} \\

\hline
\multirow{4}{*}{\textbf{HH-RLHF}}\\
\multirow{4}{*}{\textbf{(numerator)}}
& SFT $\Rightarrow$ SFT (2491)
& 1225 & 545 & 208 & 43 & 7 \\

& SFT $\Rightarrow$ RLAIF (2802)
& 1549 & 948 & 482 & 180 & 33 \\

& RLHF $\Rightarrow$ SFT (2500)
& 1113 & 513 & 193 & 44 & 7 \\

& RLHF $\Rightarrow$ RLAIF (2722)
& \textbf{1806} & \textbf{1083} & \textbf{620} & \textbf{298} & \textbf{74} \\

\hline
\end{tabular}}
\caption{Deception (broad) scores and counts across pipelines for different confidence thresholds $\tau$. 
Each column corresponds to a different threshold, illustrating how deception varies with confidence calibration. 
Across both datasets, deception is highly sensitive to $\tau$, with RLHF $\Rightarrow$ RLAIF on PKU-SafeRLHF exhibiting consistently elevated deception across all thresholds.}
\label{tab:broad-multitau}
\end{table*}

\section{Binary-indicator analysis}

As a robustness check, we also compute the bias--variance--covariance terms using binary correctness indicators rather than continuous confidence scores. In this variant, each model prediction is converted to a 0/1 indicator according to whether it assigns the higher preference to the human-preferred response. This experiment directly addresses the possible concern that the theoretical quantities are reward-valued, while our main empirical analysis relies on confidence-score proxies. However, we emphasize that this binary substitution removes information about confidence magnitude and compresses model behavior into thresholded correctness. As a result, the resulting variance terms are largely governed by Bernoulli variability and can saturate near their maximum value, which weakens the connection to the squared-loss motivation of the decomposition. For this reason, we report these results only as a sensitivity analysis and use the continuous confidence-score decomposition as the primary diagnostic throughout the paper.

\begin{table*}[h]
\centering
\resizebox{\textwidth}{!}{
\begin{tabular}{c c c c c c}
\hline
&&\textbf{$\boldsymbol{\mathrm{Bias}^2(\hat{r}^{\pi^w})}$}&\textbf{$\boldsymbol{\mathrm{Var}(\hat{r}^{\pi^w})}$}& \textbf{$\boldsymbol{\mathrm{Cov}(r^*, \hat{r}^{\pi^w})}$}&\\ 
\textbf{Dataset} & \textbf{W$\boldsymbol{\Rightarrow}$S Pipeline}& \textbf{$\boldsymbol{\mathrm{Bias}^2(\hat{r}^{\pi^s} \mid \hat{r}^{\pi^w})}$}&\textbf{$\boldsymbol{\mathrm{Var}(\hat{r}^{\pi^s})}$}& \textbf{$\boldsymbol{\mathrm{Cov}(\hat{r}^{\pi^s}, \hat{r}^{\pi^w})}$}&\textbf{$\boldsymbol{\mathcal{R}_{w2s}(\pi^s \mid \pi^w)}$}\\ 
& & \textbf{$\boldsymbol{--}$}&\textbf{$\boldsymbol{\mathrm{Var}(r^*)}$}& \textbf{$\boldsymbol{--}$}&\\
\hline

\multirow{12}{*}{\rotatebox[origin=c]{90}{\textbf{PKU-SafeRLHF}}}

& SFT $\Rightarrow$ SFT
& $4.84\times10^{-7}$ & $0.24998$ & $-0.00370$\\
&
& $1.78\times10^{-8}$ & $0.24998$ & $0.24992$ & $\underline{0.52035}$\\
&
& -- & $0.24644$ & --\\

& SFT $\Rightarrow$ RLAIF
& $4.84\times10^{-7}$ & $0.24998$ & $-0.00370$\\
&
& $3.25\times10^{-6}$ & $0.24999$ & $0.24522$ &$0.65203$ \\
&
& -- & $0.24644$ & --\\

& RLHF $\Rightarrow$ SFT
& $1.69\times10^{-5}$ & $0.24989$ & $0.00196$ \\
&
& $2.23\times10^{-4}$ & $0.24998$ & $0.24270$ & $0.68592$\\
&
& -- & $0.24644$ & --\\

& RLHF $\Rightarrow$ RLAIF
& $1.69\times10^{-5}$ & $0.24989$ & $0.00196$\\
&
& $2.25\times10^{-4}$ & $0.24998$ & $0.24267$ & $\boldsymbol{0.68637}$\\
&
& -- & $0.25489$ & --\\

\hline

\multirow{12}{*}{\rotatebox[origin=c]{90}{\textbf{HH-RLHF}}}

& SFT $\Rightarrow$ SFT
& $6.15\times10^{-5}$ & $0.24999$ & $0.00100$ \\
&
& $4.58\times10^{-9}$ & $0.24999$ & $0.24995$ & $\underline{0.49848}$ \\
&
& -- & $0.25489$ & --\\

& SFT $\Rightarrow$ RLAIF
& $6.15\times10^{-5}$ & $0.24999$ & $0.00100$ \\
&
& $4.58\times10^{-9}$ & $0.24999$ & $0.24995$ & $\underline{0.49848}$ \\
&
& -- & $0.23888$ & --\\

& RLHF $\Rightarrow$ SFT
& $5.84\times10^{-5}$ & $0.24999$ & $0.00103$ \\
&
& $1.83\times10^{-8}$ & $0.24999$ & $0.24982$ & $\boldsymbol{0.51297}$\\
&
& -- & $0.23888$ & --\\

& RLHF $\Rightarrow$ RLAIF
& $5.84\times10^{-5}$ & $0.24999$ & $0.00103$ \\
&
& $1.83\times10^{-8}$ & $0.24999$ & $0.24982$ & $\boldsymbol{0.51297}$\\
&
& -- & $0.23915$ & --\\

\hline
\end{tabular}}
\caption{Sensitivity analysis using binary correctness indicators for the bias--variance--covariance decomposition. 
Unlike the main confidence-based analysis, this version thresholds each model's preference into a 0/1 correctness indicator. 
The table is included for completeness and to examine the effect of replacing continuous confidence scores with binary reward-valued proxies. 
Because thresholding removes confidence magnitude, the resulting variance and covariance terms should be interpreted cautiously and are not used as the primary evidence for our main conclusions.}
\label{tab:bias-var-cov-binary}
\end{table*}

\section{Confidence Standard Deviation Analysis}
Table~\ref{tab:conf-avg-std} shows the standard deviation of confidence (equation~\ref{eq:conf-score-supp}) across all 5000 evaluation examples for a given model. This is tau-independent — it describes the overall shape of the confidence distribution. Low std (~0.12) means scores cluster near 0.5 (model is indecisive on most examples). High std (~0.29) means scores spread toward 0 and 1 (model has strong opinions, right or wrong).
        \begin{equation}\label{eq:conf-score-supp}
        \mathcal{C}^{\pi}(x) = \sigma(\ell_{\pi}(a^1|x) - \ell_{\pi}(a^2|x)),
    \end{equation}
\begin{table}[H]
\centering
\resizebox{\columnwidth}{!}{
\begin{tabular}{c c c c c}
\hline
\textbf{Dataset} & \textbf{W$\boldsymbol{\Rightarrow}$S Pipeline}& \textbf{Strong Avg Conf}&\textbf{Strong Std}&\textbf{Weak Std}\\ 
\hline

\textbf{PKU-SafeRLHF}

& SFT $\Rightarrow$ SFT & \Large{0.521} & \Large{0.291} & \Large{0.129}\\

& SFT $\Rightarrow$ RLAIF & \Large{0.503} & \Large{0.116} & \Large{0.129}\\

& RLHF $\Rightarrow$ SFT & \Large{0.501} & \Large{0.127} & \Large{0.123}\\

& RLHF $\Rightarrow$ RLAIF & \Large{0.502} & \Large{0.117} & \Large{0.123}\\

\hline

\textbf{HH-RLHF}

& SFT $\Rightarrow$ SFT & \Large{0.475} & \Large{0.227} & \Large{0.129}\\

& SFT $\Rightarrow$ RLAIF & \Large{0.503} & \Large{0.165} & \Large{0.129}\\

& RLHF $\Rightarrow$ SFT & \Large{0.476} & \Large{0.207} & \Large{0.123}\\

& RLHF $\Rightarrow$ RLAIF & \Large{0.503} & \Large{0.165} & \Large{0.123}\\

\hline
\end{tabular}}
\caption{Standard deviation of confidence across all 5000 evaluation examples for a given model.}
\label{tab:conf-avg-std}
\end{table}

This dataset-level confidence standard deviation is distinct from the strong-model variance $\boldsymbol{\mathrm{Var}(\hat{r}^{\pi^s})}$ reported in Table 1 of the main paper and Table 1 of this supplementary material, which is computed through the BVC estimation procedure over repeated weak-policy relabeling trials; the two quantities therefore measure different sources of variation and are not directly comparable.

    \section{Dataset Details}
        \begin{itemize}

        \item \textbf{HH-RLHF}~\cite{bai2022training} consists of helpful and harmless subsets, along with red-teaming data. We select 15{,}000 samples from the harmless subset and partition them into three disjoint splits of 5{,}000 samples each, used for weak teacher training, strong student training, and held-out evaluation, respectively.
    
        \item \textbf{PKU-SafeRLHF}~\cite{dai2024safe} contains paired responses annotated for safety and helpfulness. We select 15{,}000 samples by choosing the safer response for each prompt and divide them into three splits of 5{,}000 samples for our training and evaluation pipeline.
    \end{itemize}

    \section{Evaluation Details}

    \subsection{Estimating Bias, Variance, and Covariance from Confidence Scores}
\label{sec:acc_bias_var}

Our theoretical analysis is stated in terms of reward-valued quantities. However, explicit reward-model outputs are not available in all pipelines, particularly in SFT$\Rightarrow$SFT, where the model only defines a policy over responses rather than a separate reward function. To enable a uniform empirical comparison across all weak-to-strong pipelines, we therefore use the pairwise confidence score in Eq.~\ref{eq:conf-score-supp} as a continuous proxy for reward-like preference strength. This proxy is available for every model, preserves magnitude information that binary accuracies discard, and provides a more faithful basis for estimating second-order quantities such as variance and covariance. We estimate expectations over $a\sim \pi^w$ by repeatedly sampling completions from the weak policy and re-running the relabeling/evaluation pipeline.

\paragraph{Step 1: Estimating the mean reward terms.}
For a given prompt $x$, we estimate the conditional reward moments induced by the weak policy. Specifically, we run the weak relabeling procedure for $\mathrm{N}$ i.i.d.\ trials (here $N=3$) while keeping the prompt fixed.
In trial $n$, the weak policy $\pi^w$ produces an output (action) $a^{(n)}\sim \pi^w(\cdot\mid x)$. Thus, the randomness in this step comes from the weak policy’s action distribution for the same prompt, allowing us to isolate how weak-policy sampling affects the BVC terms. We then define three empirical preference confidence indicators:

\begin{itemize}
    \item \textbf{Ground-truth preference confidence under weak actions.}
    Let $r^{*(n)} \triangleq \mathcal{C}^{\pi^w}_{gt}(x^{(n)}) \in [0,1]$ denote the weak model's confidence score evaluated on the human-labeled preference pair for prompt $x^{(n)}$  in $\mathcal{D}_2$ . Concretely, the pair $(a^1, a^2)$ is ordered such that $a^1 \succ a^2$ according to the ground-truth human label, and $\mathcal{C}^{\pi^w}_{gt}(x^{(n)})$ is computed using Eq.~\eqref{eq:conf-score-supp}. This serves as a confidence-based empirical proxy for the target reward term.

    \item \textbf{Weak preference confidence on weak-labeled data.}
    Let $\hat r^{\pi^w,(n)} \triangleq \mathcal{C}^{\pi^w}_{w}(x^{(n)}) \in [0,1]$ denote the weak model's confidence score evaluated on the weak-labeled preference pair for the same prompt  in $\mathcal{D}_2$. Here, the pair $(a^1, a^2)$ is ordered according to the weak model's pseudo-label, and the confidence is again computed using Eq.~\eqref{eq:conf-score-supp}.

    \item \textbf{Strong preference confidence on weak-labeled data.}
    Let $\hat r^{\pi^s,(n)} \triangleq \mathcal{C}^{\pi^s}_{w}(x^{(n)}) \in [0,1]$ denote the strong model's confidence score evaluated on the same weak-labeled preference pair  in $\mathcal{D}_2$ , computed analogously using Eq.~\eqref{eq:conf-score-supp}.
\end{itemize}

Using these confidence proxies, we estimate the mean reward terms by sample averages:
\begin{align}
\bar r^* &\approx \frac{1}{\mathrm{N}}\sum_{n=1}^{\mathrm{N}} r^{*(n)}, \\
\bar r^{\pi^w} &\approx \frac{1}{\mathrm{N}}\sum_{n=1}^{\mathrm{N}} \hat r^{\pi^w,(n)}, \\
\bar r^{\pi^s} &\approx \frac{1}{\mathrm{N}}\sum_{n=1}^{\mathrm{N}} \hat r^{\pi^s,(n)} .
\end{align}
All three expectations are taken with respect to $a\sim\pi^w$; empirically, this corresponds to
using the weak policy as the sampling distribution and re-running the corresponding evaluation
procedure $\mathrm{N}$ times (here $N=3$).

\paragraph{Step 2: Estimating bias, variance, and covariance.}
Given the empirical means above, we compute the squared bias terms as:
\begin{align}
\mathrm{Bias}^2(\hat r^{\pi^w}) &\triangleq (\bar r^{\pi^w}-\bar r^*)^2, \\
\mathrm{Bias}^2(\hat r^{\pi^s}\mid \hat r^{\pi^w}) &\triangleq (\bar r^{\pi^s}-\bar r^{\pi^w})^2.
\end{align}

For the remaining second-order terms, we estimate variances/covariances by sample moments under the
same weak-action distribution   in $\mathcal{D}_2$ :
\begin{align}
\mathrm{Var}(\hat r^{\pi^w})
&\approx \frac{1}{\mathrm{N}}\sum_{n=1}^{\mathrm{N}}\big(\hat r^{\pi^w,(n)}-\bar r^{\pi^w}\big)^2, \\
\mathrm{Var}(r^*)
&\approx \frac{1}{\mathrm{N}}\sum_{n=1}^{\mathrm{N}}\big(r^{*(n)}-\bar r^*\big)^2, \\
\mathrm{Cov}(r^*,\hat r^{\pi^w})
&\approx \frac{1}{\mathrm{N}}\sum_{n=1}^{\mathrm{N}}\big(r^{*(n)}-\bar r^*\big)\big(\hat r^{\pi^w,(n)}-\bar r^{\pi^w}\big), \\
\mathrm{Var}(\hat r^{\pi^s})
&\approx \frac{1}{\mathrm{N}}\sum_{n=1}^{\mathrm{N}}\big(\hat r^{\pi^s,(n)}-\bar r^{\pi^s}\big)^2, \\
\mathrm{Cov}(\hat r^{\pi^s},\hat r^{\pi^w})
&\approx \frac{1}{\mathrm{N}}\sum_{n=1}^{\mathrm{N}}\big(\hat r^{\pi^s,(n)}-\bar r^{\pi^s}\big)\big(\hat r^{\pi^w,(n)}-\bar r^{\pi^w}\big).
\end{align}

\paragraph{Reporting the misfit bound (including covariances).}
Finally, we report corresponding bound on the misfit that depends on the covariance terms.
In our setting, the bound is computed by substituting the empirical estimates of
$\mathrm{Bias}^2(\hat r^{\pi^w})$, $\mathrm{Bias}^2(\hat r^{\pi^s}\mid \hat r^{\pi^w})$,
$\mathrm{Var}(r^*)$, $\mathrm{Var}(\hat r^{\pi^w})$, $\mathrm{Var}(\hat r^{\pi^s})$,
$\mathrm{Cov}(r^*,\hat r^{\pi^w})$, and $\mathrm{Cov}(\hat r^{\pi^s},\hat r^{\pi^w})$
into the misfit inequality.

\paragraph{Held-out evaluation for deception metrics.}
All deception evaluations are performed on the held-out subset $\mathcal{D}_3$. This subset is not used for training the weak model, weak relabeling, or training the strong model. Instead, $\mathcal{D}_3$ contains unseen human-annotated preference examples and is used only for evaluation. For each example in $\mathcal{D}_3$, we compute the weak and strong confidence scores with respect to the ground-truth preference label and then evaluate blind-spot deception. This ensures that the reported deception scores measure out-of-sample weak-to-strong failures rather than behavior on examples used during training or weak relabeling.

\section{Standard RLHF Background}
\label{app:standard_rlhf}

    For completeness, we summarize the standard RLHF notation~\citep{xiong2024iterative} used in the main text. We formulate the RLHF process as aligning a large language model (LLM) to take a prompt $x\in\mathcal{X}$, and output a response $a = [t_1, t_2, \dots]$, where $t_i$ is the $i$-th token generated by the LLM. From a reinforcement learning point of view we call $\mathcal{X}$ the state-space of a contextual bandit with action-space $\mathcal{A}$. In RLHF a ground truth reward function $r^*(x,a)$ maps from the state-action space of this bandit to a chosen-rejected 1 or 0 reward, more formally we have:

    \begin{equation}
    \label{eq:rew_func}
        r^*: \mathcal{X}\times \mathcal{A}\rightarrow [0, 1].
    \end{equation}

    \subsection{Reward Modeling}
    
    Typical RLHF datasets consist of tuples $(x, a^1, a^2, y)$ with $y$ being the preference signal defined in~\ref{eq:preference}:

    \begin{equation}
    y=
    \begin{cases}
    1\;\;\;\;a^1 \succ a^2\\
    0\;\;\;\;a^1 \prec a^2,
    \end{cases}
    \label{eq:preference}
    \end{equation}

    \noindent where $a^1 \succ a^2$ means response $a^1$ is preferred over response $a^2$.     

    To model the the reward function defined in~\ref{eq:rew_func} we follow the \cite{bradley1952rank} model:

    \begin{equation}
    \label{eq:Bradley_Terry}
    \begin{aligned}
        &\mathbb{P}(a^1 \succ a^2|x,a^1, a^2)\\ 
        &= \frac{\exp(r^*(x,a^1))}{\exp(r^*(x,a^1))+\exp(r^*(x,a^2))}\\ 
        &= \sigma(r^*(x,a^1)-r^*(x,a^2)),
    \end{aligned}
    \end{equation}

    \noindent where $\sigma(z) = \frac{1}{1+exp(-z)}$ is the sigmoid function. 

    \subsection{Policy Training}

    In aligning the LLMs using RLHF, we first have a pretrained model fine-tuned on instruction following tasks denoted as $\pi_0$. The goal is to align the LLM $\pi$ to a set of prompts taken from distribution $x\sim d_0$, while keeping it close to the $\pi_0$. Therefore we maximize the objective~\ref{eq:rlhf_objective}.

    \begin{equation}\label{eq:rlhf_objective}
    \begin{aligned}
            &J(\pi)\\ 
            &=\mathbb{E}_{x\sim d_0}[\mathbb{E}_{a\sim\pi(\cdot\mid x)}[r^*(x,a)] - \eta D_{KL}(\pi(\cdot\mid x)||\pi_0(\cdot\mid x))],
    \end{aligned}
    \end{equation}

    \noindent where $\eta > 0$ is coefficient for the KL-penalty term.
    After this process the aligned LLM $\pi$ is a mapping from the state-space $\mathcal{X}$ to a distribution over the action-space $\mathcal{A}$. These equations are standard and are included only to define notation.

    \section{Implementation Details}

    Our weak-to-strong framework consists of a single weak model and a single strong model. We use Llama-3.2-3B-Instruct to train the weak model and Llama-3.1-8B-Instruct to train the strong model.

    During both training and inference, we adopt a simple instruction–response format for single-turn conversations, where user inputs are prefixed with \texttt{User:} and model outputs with \texttt{Assistant:}. No multi-turn context or additional role conditioning beyond these prefixes is used. We use a learning rate of $2\times10^{-5}$ for reward modeling, $1\times10^{-5}$ for policy optimization, and $1\times10^{-5}$ for supervised fine-tuning (SFT). Batch sizes are set to 4 for reward modeling, 8 for policy training, and 2 for SFT. Hyperparameters are selected based on pilot experiments.

    We do not employ LoRA or other parameter-efficient fine-tuning techniques. All experiments are conducted on 1 to 4 NVIDIA H100 GPUs with 80\,GB of memory. 

    After training the weak model, we use it to relabel preference data by prompting it with a fixed comparison template together with the original dataset text. For each instance, the model is asked to select the preferred option and reject the alternative, producing a binary preference label. This procedure converts the original samples into weakly supervised preference pairs, which are subsequently used for training downstream components. 
    \section{Reproducibility Statement}
    All reported results are obtained by averaging over three independent runs with different random seeds $\{42,  52, 62\}$. To support reproducibility and facilitate further analysis, we release the full codebase used in our experiments at \url{--}.

\section{Limitations}

Our empirical study has several important limitations. First, the experiments are restricted to Llama-family models, so the observed relationships may not transfer directly to other architectures or model families. Second, our evaluation is conducted on pairwise preference datasets centered on safety and harmlessness, which leaves open whether the same patterns hold in domains such as reasoning~\citep{luowizardmath,yang2024weak}, coding~\citep{luowizardcoder,liu2023your}, or multi-objective alignment. Third, our correlation-based findings are based on only eight pipeline/dataset combinations, so they should be interpreted as exploratory empirical evidence rather than definitive statistical laws. More broadly, the results in this paper should be viewed as a diagnostic characterization of weak-to-strong failure modes in the evaluated settings, not as a universal account of deception in all alignment pipelines. The deception metric is an operational diagnostic for identifying high-risk failure regions where weak-model signal is absent, rather than a definitive proof of strategic intent.

A further limitation is that our study does not test closed-loop mitigation. Although strong-model variance is strongly associated with blind-spot deception in our experiments, we do not show that using this signal during training reduces deception. For example, we do not use $\mathrm{Var}(\hat r^{\pi^s})$ as an explicit regularizer, early-stopping criterion, checkpoint-selection rule, or data-selection signal. Therefore, the practical value of strong-model variance should be interpreted as a candidate early-warning signal rather than a demonstrated intervention.

Future work should test whether this diagnostic signal can be used actively to reduce weak-to-strong failures. Promising directions include penalizing excessive confidence dispersion during training, selecting checkpoints with lower held-out strong-model variance, prioritizing data where weak-model uncertainty and strong-model confidence disagree, or filtering weak labels in regions associated with high blind-spot risk. These interventions would provide the closed-loop validation needed to determine whether the diagnostic signal identified here can be converted into a practical mitigation strategy.

\end{document}